%% file: root.tex
\documentclass[10pt,letterpaper,conference]{ieeeconf}

\usepackage{latex_resources/saurav_macros}
\usepackage{latex_resources/shorthands_minimal}
\usepackage[dvipsnames]{xcolor}
\usepackage{siunitx}
\usepackage[caption=false,font=footnotesize]{subfig}

\pgfplotsset{compat=1.18}

\usepackage{cite}

\newcommand{\normtwo}[1]{\left\lVert#1\right\rVert_2}

\IEEEoverridecommandlockouts                              %
\makeatletter
\long\def\@makecaption#1#2{%
\ifx\@captype\@IEEEtablestring%
\setbox\@tempboxa\hbox{\footnotesize #1:~~ #2}%
\ifdim \wd\@tempboxa >\hsize%
\setbox\@tempboxa\hbox{\footnotesize #1:~~ }%
\parbox[t]{\hsize}{\footnotesize \noindent\unhbox\@tempboxa#2}%
\else \hbox to\hsize{\footnotesize\hfil\box\@tempboxa\hfil}\fi%
\@IEEEtablecaptionsepspace%
\else
\@IEEEfigurecaptionsepspace%
\setbox\@tempboxa\hbox{\footnotesize #1.~~ #2}%
\ifdim \wd\@tempboxa >\hsize%
\setbox\@tempboxa\hbox{\footnotesize #1.~~ }%
\parbox[t]{\hsize}{\footnotesize \noindent\unhbox\@tempboxa#2}%
\else%
\ifcenterfigcaptions \hbox to\hsize{\footnotesize\hfil\box\@tempboxa\hfil}%
\else \hbox to\hsize{\footnotesize\box\@tempboxa\hfil}%
\fi\fi\fi}
\makeatother

\title{\vspace{6mm} \LARGE \bf Controlling Collectives of AI Agents in \\ Reasoning Space with Spatial Transformers}

\author{Frederic Vatnsdal, Roshan Gopal, Romina Garcia Camargo, Vijay Kumar, and Alejandro Ribeiro%
\thanks{\raggedright \mbox{F.~Vatnsdal}, \mbox{R.~Garcia Camargo}, and \mbox{A.~Ribeiro} are with the Department of Electrical and Systems Engineering, University of Pennsylvania, USA. \mbox{R.~Gopal} is with the School of Arts and Sciences, University of Pennsylvania, USA. \mbox{V.~Kumar} is with the GRASP Laboratory, University of Pennsylvania, USA. Emails: {\tt\{vatnsdal, rominag, kumar, aribeiro\}@seas.upenn.edu}, {\tt rgopal@sas.upenn.edu}. This work was supported by grant ARL DCIST CRA W911NF-17-2-0181.}%
}

\begin{document}
\bstctlcite{IEEEexample:BSTcontrol}
\maketitle

\begin{abstract}
Large Language Models (LLMs) introduce an exciting new paradigm for planning and navigation in robotics, but fail on even simple multi-robot tasks as team sizes grow.
We propose COMPASS, a scalable, decentralized multi-robot architecture for controlling large collectives of agentic robots with reasoning space feedback control.
Feedback is generated locally on each robot by a spatial transformer which aggregates multi-hop messages across the fleet into a learned feedback token.
Our experiments find that collectives of language models demonstrate performance gains from structured diversity of the input command, which can cancel biases; an advantage that is held across scale.
Compared against a centralized frontier LLM policy and a language-only communication ablation, we find that the coupled design of COMPASS decisively produces cohesive flocking formations that accurately fly the commanded intent.
We show that reasoning feedback works best when composed with a compact learned token. 
Our ablations show that hand engineered feedback with raw state appearing in the language channel obliterates cohesion.
COMPASS generalizes zero-shot to unseen instructions of ambiguous meaning while commanding flocks up to $16\times$ its training scale, flying up to $1024$ robots under natural language commands.
\end{abstract}

\section{Introduction}
\label{sec:intro}
\input{sections/01_intro}

\section{Reasoning-Space Feedback Control of Collective Agents}
\label{sec:lang_models}
\input{sections/02_lang_models}

\section{Training for Consensus}
\label{sec:training}
\input{sections/05_implementation}

\section{Results}
\label{sec:results}
\input{sections/04_results}

\section{Conclusion}
\label{sec:conclusion}
\input{sections/06_conclusion}

\bibliographystyle{IEEEtran}
\bibliography{IEEEabrv,MALLM-zotero, MALLM-manual}  %
\end{document}

%% file: sections/01_intro.tex
\begin{figure}[ht!]
    \centering
    \includegraphics[width=\linewidth]{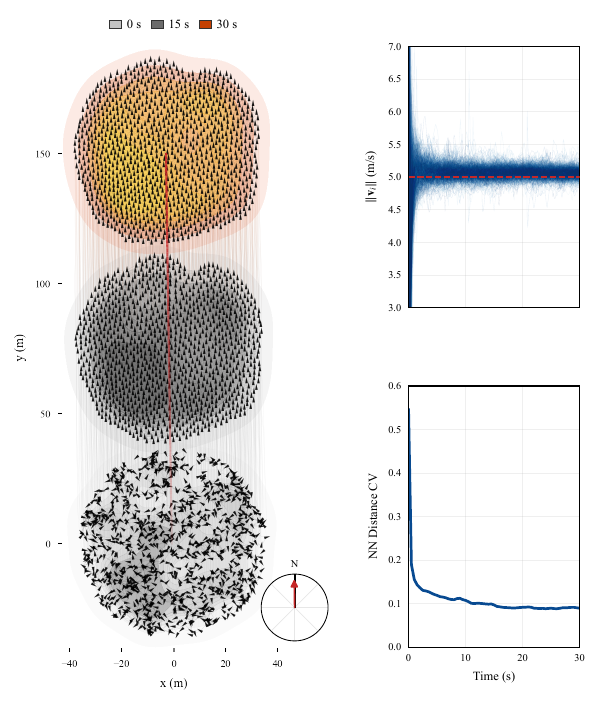}
    \caption{\textbf{Left:} ``Fly toward twelve o'clock'' -- the $N=1024$ decentralized swarm individually decodes the language instruction and mobilizes north-bound at $\qty{5}{\meter\per\second}$, shown at $0$, $15$, and $\qty{30}{\second}$ (agents black, communication graph at $r_c = \qty{4}{\meter}$).
    The center of mass is drawn as a red throughline.
    \textbf{Top Right:} the speed of every agent converges to the commanded $\qty{5}{\meter\per\second}$ over the rollout.
    \textbf{Bottom Right:} the coefficient of variation of the robots' nearest-neighbor distances 
    converges as the flock settles into an even lattice.
}
    \label{fig:compass_headline}
\end{figure}
Embedding language models (LMs) into robotic planning and control frameworks harnesses general-purpose reasoning for embodied tasks, granting unprecedented zero-shot capabilities \cite{kwon2024language, ahnCanNotSay, varley2024embodiedai, dorbala2024canan, singh2025malmm}.
The collaborative feedback control of large-scale collectives of embodied agents could enable robust reasoning over challenging tasks, enabling sophisticated multi-robot reasoning.
We present COmmanding Multi-robot Policies with Attention and Semantic Steering (COMPASS); \fgref{fig:compass_headline}.
COMPASS is a decentralized, feedback control architecture operating in reasoning space that integrates spatial transformers in the loop to infuse collective fleet embeddings into the feedback tokens injested by each robot's LM.
COMPASS out-scales and outperforms approaches that rely only on the language-channel for coordination.

The methods for controlling embodied language models at scale remain underdeveloped in the literature \cite{Chen2023ScalableMC}.
Providing a language model with overwhelming knowledge about teammates can easily corrupt reasoning over the original task, hinting that new methods must be developed to leverage reasoning at scale \cite{shi2023llmsdistracted, guo2026embodied}.
Our empirical study of frontier models reveals that language models struggle with spatial awareness and reasoning over the states of teammates while simultaneously completing a navigation task; a limitation that COMPASS addresses directly.

Language commanded flocking is our chosen test-bed for examining how collaborative reasoning feedback loops to both steer a fleet toward a common objective while maintaining cohesive formations.
COMPASS builds on the perception--action-communication literatures, with learned communication backbones Graph Neural Networks (GNNs) \cite{agarwalLPACLearnablePerceptionActionCommunication2024, tolstayaLearningDecentralizedControllers2020a} and spatial transformers \cite{owerko2025mast} that provide the scalability that language models lack on their own.
We make the following contributions:
\begin{itemize}
    \item \textbf{Scale:} zero-shot generalization to $16\times$ the training team size on kilo-scale flocks.
    \item \textbf{Structured Diversity:} COMPASS has the unique ability to spread multiple phrasings of the same task across the fleet, so that a minority running an easier phrasing can raise the accuracy of the collective.
    \item \textbf{Generalization:} COMPASS taps into the latent dimension of language models with a learned query-only interface that is able to ground unseen directions to executable intents.
\end{itemize}

\subsection{Related Work}
\textbf{Large Language Models (LLM) for Mission Planning} \: Integrated into control frameworks, LLMs enable zero-shot execution across robotic navigation and manipulation tasks \cite{kwon2024language, ahnCanNotSay, varley2024embodiedai}. Kwon et al. \cite{kwon2024language} demonstrate that an LLM can generate low-level end-effector trajectories using only off-the-shelf vision modules. SayCan \cite{ahnCanNotSay} demonstrated that LLMs are effective high level mission planners due to their rich semantic reasoning. Nonetheless, the models require grounding to work in robotics tasks that have more stringent physical constraints and metrics of success, often achieved through vision or human interaction \cite{gao2024physically, wang2024icantell}. Approaches for online planning with LLMs incorporate feedback from the environment as well as previously generated plans to influence subsequent generations \cite{ravichandranSPINEOnlineSemantic2025, pmlr-v205-huang23c, ravichandranDistillingOndeviceLanguage2025}.
PRISM \cite{ravichandranDistillingOndeviceLanguage2025} distilled online semantic planning into SLMs that run on robot hardware.
Building on this approach, Ro-SLM \cite{wang-etal-2026-ro} optimizes on-device SLMs by employing the LLM as a reward function to guide fine-tuning on synthesized task-planning data.

\textbf{Coordination of LLMs for Multi-Robot Tasks} \: Recent works study the capabilities of LLMs for coordinating multi-robot teams with centralized \cite{kannanSMARTLLMSmartMultiAgent2024, Chen2023ScalableMC, liu2025coherent} or decentralized approaches \cite{liLLMFlockDecentralizedMultiRobot2025, mandiRoCoDialecticMultiRobot2024, venkateshZeroCAPZeroShotMultiRobot2025, yang2026decentralised}. Decentralized consensus mechanisms generate robust team plans that avoid instability and inconsistent behaviors. 
LLM-Flock \cite{liLLMFlockDecentralizedMultiRobot2025} addresses formation control of robots, each endowed with an independent LLM generating plans locally.
RoCo \cite{mandiRoCoDialecticMultiRobot2024} uses a dialectic style of communication between LLM agents that govern individual robots, with focus on multi-robot manipulation that supports human interaction.
ZeroCAP \cite{venkateshZeroCAPZeroShotMultiRobot2025} demonstrated that a VLM-LLM combined architecture enables reasoning about spatial tasks with multi-robot teams such as infilling and encircling objects.

%% file: sections/02_lang_models.tex
\begin{figure*}[ht]
    \centering
    \includegraphics[width=\linewidth]{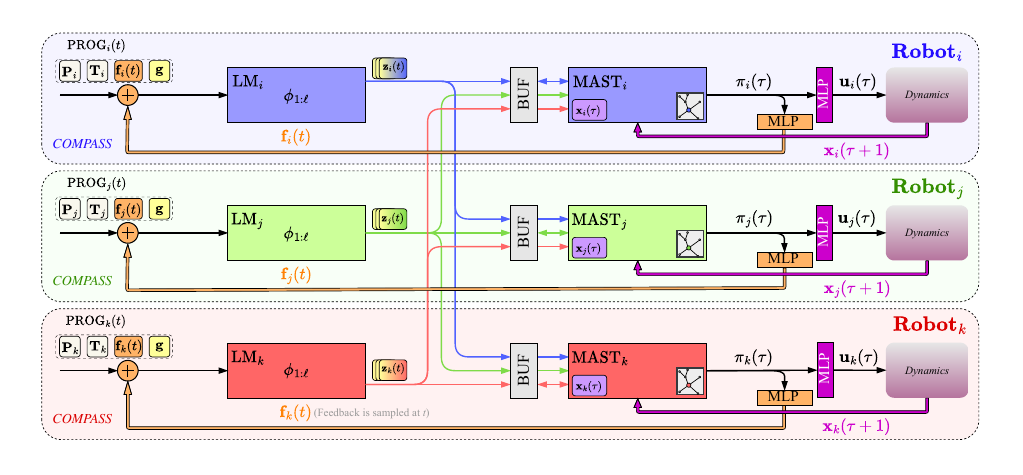}
    \caption{
        COMPASS uses local LMs to update the reasoning state $\bf{z}(t)$ from both the command intent and the learned feedback token (orange) produced by MAST from neighborhood information.
        COMPASS is deployed identically on each robot despite being trained centralized with spatial masking.
        Neighboring robots exchange reasoning over the communication network and store received communications in buffers.
        The same MAST output used to produce feedback is also projected into the action space for execution of low-level control (purple).
        MAST incorporates low-level state feedback into its input features.
    }
    \label{fig:arch}
\end{figure*}
COMPASS pilots a team of $N$ networked robots with physical state at control step $\tau$ $\bfX(\tau) = [\bfx_1(\tau), \cdots, \bfx_N(\tau)]^\top \in \reals^{N \times d_x}$ and actions $\bfU(\tau) = [\bfu_1(\tau), \cdots, \bfu_N(\tau)]^\top \in \reals^{N \times d_u}$.
Alongside this fast physical system, each robot $i$ carries an LM and a persistent \emph{program} of embedded tokens,
\begin{equation}\label{eq:program}
    \mathrm{PROG}_i(t) = \left[ \bfP_i \, \Vert \, \bfT_i \, \Vert \, \bff_i(t) \, \Vert \, \bfg \right],
\end{equation}
that is comprised of the operator prompt $\bfP$, a thinking trace $\bfT$, a feedback token $\bff$, and a learned goal token $\bfg$, all in the model's embedding space $\reals^{d_{\mathrm{LM}}}$.
COMPASS is therefore formed by the coupling of a physical dynamical system with latent reasoning dynamics, optionally evolving on separate timescales; \fgref{fig:arch}.
The fleet's state $\bfX(\tau)$ is advanced by a spatial transformer (\secref{sec:spatialtransformer}), while a latent \emph{reasoning state} $\bfz_i(t)$ updates at the slower reasoning rate.
Reasoning is re-derived from the program each time the feedback token is refreshed with a projection of the fleet state (\secref{sec:feedback}).
The prompt $\bfP$, e.g., ``vector two o'clock'', is updated when the operator redirects the fleet, triggering a regeneration of $\bfT$.

\subsection{Reasoning in the Loop}
\label{sec:feedback}
COMPASS reasons through both \emph{thinking} in language space at generation time and \emph{latent} reasoning at control time.
The generic update of the reasoning state is
\begin{equation}\label{eq:reasoning_update}
  \bfz_i(t+1) = \phi_{1:\ell}\big(\, \bfP_i \Vert \bfT_i \Vert \psi(\bfpi_i(t)) \Vert \bfg \,\big),
\end{equation}
where $\phi_{1:\ell}$ denotes the first $\ell$ layers of the LM, $\psi$ is a learned adapter that embeds the output of the downstream controller $\bfpi_i(t)$ (\secref{sec:spatialtransformer}) as the feedback token $\bff_i(t)$.
Equation \eqref{eq:reasoning_update} defines a family of collective reasoning space controllers.

\subsubsection{Thinking}
Each robot generates its own trace conditioned on the prompt, one token at a time using all layers and the language head,
\begin{equation}
\bfT_i(\tau + 1) = \bfT_i(\tau) \,\Vert\, \phi_{\mathrm{gen}}\big( \bfP_i \Vert \bfT_i(\tau) \big),
\end{equation}
until a fixed thinking length is reached.
Traces are sampled by each individual robot and $N$ robots produce $N$ distinct readings of the same instruction; diversity that the fleet must reconcile.
Once complete, a trace is frozen until the next prompt change and KV-cached.
Every subsequent latent update in \eqref{eq:reasoning_update} reuses it, saving computation cost.

\subsubsection{Latent}
The reasoning state $\bfz_i(t)$ is the hidden state at the goal token's position after a forward pass of only $\ell$ of the LM's $L$ layers over the program.
Because of causal masking and their terminal position, the feedback and goal tokens query the entire program while no language token attends to them.
This simplifies learning as the loop tokens $\bff$ and $\bfg$ need not lie on the language manifold.
The extraction depth $\ell$ is a design parameter; empirically, vague commands require deeper extraction but performance does not always improve monotonically.
Stacking the latent states of the fleet gives $\bfZ(t) = [\bfz_1(t), \cdots, \bfz_N(t)]^\top$, the collective reasoning state of the system.

\subsection{Spatial Transformer}
\label{sec:spatialtransformer}
A multi-agent spatial transformer (MAST) \cite{owerko2025mast} turns the fleet's reasoning states into actions, diffusing them hop-by-hop over the communication graph using neighbor communication only.
The input to MAST is constructed by concatenating the reasoning state with the observed state, $\bfh_i^{(0)} = \mathrm{MLP}\big( [\, \bfz_i \Vert \bfx_i \,] \big)$, where $\bfh_i^{(0)} \in \reals^{d_{\mathrm{MAST}}}$.
Absolute position never enters MAST; geometry enters attention only through relative encodings, making the policy translation invariant without breaking permutation equivariance.

2D rotary positional encoding \cite{su2023roformerenhancedtransformerrotary} injects geometry into the attention logits by splitting the query and key into $2\zeta$ channel pairs indexed by axis $a \in \{x, y\}$.
Choosing the encoding period $\lambda$, we sweep frequencies $\omega_f = 2\pi f / \lambda$ for $f = 1, \ldots, \zeta$ with $\zeta = d_{\mathrm{MAST}}/4$, and each channel pair is rotated by the matching coordinate of the robot's position $\bfp_i$,
\begin{equation}
    \tilde{\bfq}_i^{(a,f)} = R\big(\omega_f\, p_{i}^{a}\big)\, \bfq_i^{(a,f)},
    \qquad
    \tilde{\bfk}_j^{(a,f)} = R\big(\omega_f\, p_{j}^a\big)\, \bfk_j^{(a,f)},
\end{equation}
generating rotated queries and keys, $\tilde{\bfq}_i , \tilde{\bfk}_j \in \reals^2$ (dropping the $(a, f)$ notation for clarity).
Then, $R(\theta)$ is the $2 \times 2$ rotation matrix, and $\bfq_i, \bfk_i \in \reals^{d_{\mathrm{MAST}}}$ are the original agent query and key, respectively.
Since $R(\alpha)^\top R(\beta) = R(\beta - \alpha)$, the resulting logit
\begin{equation}
    \tilde{\bfq}_i^\top \tilde{\bfk}_j
    = \sum_{a,f} \bfq_i^{(a,f)\top}\, R\big(\omega_f (p_j^a - p_i^a)\big)\, \bfk_j^{(a,f)}
\end{equation}
depends on positions only through the displacement $\bfp_j - \bfp_i$ \cite{su2023roformerenhancedtransformerrotary}.
The encoding period $\lambda$ upper-bounds the resolvable geometry: too short and distant neighbors alias, too long and the encoding is inert.
Rotary encoding is applied at the first MAST layer only.

Attention is restricted to the communication graph.
With $d_{ij} = \normtwo{\bfp_j - \bfp_i}$ and mask $\calM_{ij} = 1$ iff $d_{ij} \le r_c$, per head of dimension $d_h$,
\begin{equation}
    A_{ij} = \underset{j \,:\, \calM_{ij} = 1}{\mathbf{sm}} \left( \tilde{\bfq}_i^\top \tilde{\bfk}_j \, / \sqrt{d_h} \right),
\end{equation}
so a robot attends only to teammates within communication range.
Messages are additionally conditioned on edge geometry through the normalized displacement $\bfe_{ij} = [\, \bfp_j - \bfp_i \Vert d_{ij} \,] / r_c$, for $\bfh_i^{(\ell)} = \bfh_i^{(\ell-1)} + \bfW_o \textstyle\sum_{j} A_{ij} \big( \bfW_v \bfh_j^{(\ell-1)} + \bfW_e \bfe_{ij} \big),$ followed by a pre-norm residual MLP.
Here, the learned parameters per layer and per head are, $\bfW_o, \bfW_v \in \reals^{{d_h} \times d_{h}}$ and $\bfW_e \in \reals^{d_h \times 3}$.
Each masked attention layer is exactly one round of message exchange with 1-hop neighbors, so $L_{\mathrm{MAST}}$ layers give every robot an $L_{\mathrm{MAST}}$-hop receptive field.
Therefore, due to overlapping neighborhoods, information is able to diffuse across the swarm at each step.
The communication graph's regularity is learned by MAST allowing the same parameters at $N=64$ to generalize to $N=1024$.
MAST is trained centralized but can be executed decentralized as each robot only needs to evaluate its own attention row from tokens received over the network.

The final token $\bfpi_i = \bfh_i^{(L_{\mathrm{MAST}})}$ is decoded twice to obtain feedback and action.
An MLP head produces the action $\bfu_i = \mathrm{MLP}(\bfpi_i)$ with neighbor actions predicted as byproducts but only its own is executed.
A second MLP head acts as an adapter to the LM's latent space and learns $\bff_i = \mathrm{MLP}(\bfpi_i)$, closing the reasoning loop of \eqref{eq:reasoning_update}.

%% file: sections/05_implementation.tex
 \begin{table*}[htb!]
    \centering
    \caption{Training and Evaluation Corpus}
    \label{tab:corpus}
    \begin{tabular}{llccccc}
      \toprule
      Class & Example & Prompts & Surfaces & Templates & Dirs. & Tr/Va/Te [\%] \\
      \midrule
      Plain             & ``steer North''                                & 1920 & 64  & 30 & 8   & 45/10/45 \\
      Abbreviation      & ``go N''                                 & 1920 & 64  & 30 & 8   & 45/10/45 \\
      Opposite          & ``the opposite of South''                & 1920 & 64  & 30 & 8   & 45/10/45 \\
      Map               & ``toward the top of the map''            & 1920 & 64  & 30 & 8   & 45/10/45 \\
      Clock             & ``toward twelve o'clock with twelve at the top'' & 1920 & 64 & 30 & 8 & 45/10/45 \\
      Numeric           & ``travel 0.0''                                  & 5760& 720 & 8  & 720 & 50/10/40\\
      Hold              & ``stop.''                                & 21   & 21  & -- & --  & 60/10/30 \\
      \midrule
      Vague/Conceptual  & ``all units where Santa Claus lives''    & 384  & 48  & 8  & 8   & 0/0/100 \\
      Secondary interc. & ``navigate North-northeast''            & 192  & 24  & 8  & 8   & 0/0/100 \\
      \bottomrule
    \end{tabular}
  \end{table*}
Flocking under language specified directions -- ``head East'', ``steer toward dusk'' -- is our demonstrable problem, selected for its key properties: (i) the velocities of each robot are scored against the given semantic instruction, enabling reasoning quality to be measured as heading error in degrees, (ii) the policy must balance the semantic objective with safety by avoiding collisions learned from the expert rather than scripted, (iii) global heading coherence emerges from neighbor interactions only over long horizons.
COMPASS is an incarnation of reasoning-as-policy trained for flocking where the task enters only through the expert that supervises training.
Training fits the learned parameters of the loop in \eqref{eq:reasoning_update}: the goal and feedback tokens as well as MAST; the language model is frozen.
The training data for behavior cloning is comprised of pre-generated thinking traces and centralized expert rollouts.
The seven training classes of operator prompt shown in \tabref{tab:corpus} (above the rule) generate traces with the SLM Qwen3-1.7B \cite{yang2025qwen3technicalreport}, sampled at $\text{temperature}=0.6$, $\text{top-p}=0.95$ and $\text{top-K}=20$ under a fixed multiple-choice wrapper.
The expert rollouts contain state-action pairs sampled at $dt = \qty{0.1}{\second}$ for teams of size $N=[32, 64]$, initialized uniformly on a disk of radius proportional to $\sqrt{N}$ at a target density of $0.25$ robots per $\unit{m^2}$ with random velocities and flown for $T = \qty{10}{\second}$.

Our chosen oracle is a centralized potential-field controller comprised of four terms summed per robot: (i) alignment with the fleet's mean velocity, (ii) tracking of the target velocity, (iii) cohesion toward the group centroid, (iv) collision avoidance through a repulsion term.
We train the decentralized policy to imitate this behavior from neighbor information alone.

To construct the language-action dataset, each prompt is human decoded to obtain the ground-truth intent.
Each sample in the language component is constructed through the combination of prompt \emph{surface}, i.e., the style and direction, and \emph{template}, e.g., ``go \{surface\}''.
This composite is then ingested by Qwen3-1.7B to generate $64$ thinking traces to give each robot a unique trace during training; \tabref{tab:corpus}.
The ground truth direction associated with the prompt informs the target velocity direction for the expert; for directional commands, the target speed is $\qty{5}{\meter\per\second}$.
The directions in the training set map to the 8-point rose and STOP; the exception being the angular prompt class, which sees $720$ numerically given bearings ($0.5^\circ$ increments).
We follow the aeronautical convention, North at $0^\circ$ moving clockwise to East at $90^\circ$; this matches the native assumption of the language model, measured through probing.
Therefore, this mapping is never included in the prompt.
COMPASS is trained with behavior cloning against the centralized expert, augmented by DAgger relabelling and truncated backpropagation through time for the recurrent feedback.
We remark that we leave the weights of the LM frozen, relying only on learned tokens $\bff$ and $\bfg$. 
This reduces training requirements while simultaneously leaving the optimization of the LM itself as a possible axis to better fit COMPASS to new tasks in the future.
MAST uses $L_{\mathrm{MAST}} = 4$ layers of hidden size $256$ with $8$ heads of size $d_h = 32$, a rotary encoding period of $\lambda = \qty{96}{\meter}$, and a communication radius of $r_c = \qty{4}{\meter}$.
Each robot builds the input token for MAST from its reasoning state $\bfz_i$, its own velocity $\bfv_i$,  and $\bfs_i = [\mu_{\calN_i}, v_e]$, the mean neighbor distance and velocity error against the neighborhood mean. 

We use the AdamW optimizer with the learning rate set to $10^{-4}$, weight decay $10^{-4}$ and a cosine annealing learning rate scheduler.
COMPASS is trained end-to-end for $200$ epochs on an NVIDIA RTX A5000 with $\qty{24}{\giga\byte}$ of video memory.
Evaluation is executed on a pair of NVIDIA RTX A5000 to support larger fleets of language models.

%% file: sections/04_results.tex
We evaluate COMPASS on its ability to fly cohesive flocks  up to kilo-scale team sizes (\secref{sec:res_scale}), with structured prompt diversity (\secref{sec:res_collective}), and from new directives (\secref{sec:res_lang}).
The operator directives fall on the in-distribution directions: the 8-point compass rose, numerically given bearings and Stop.
An additional 8 out-of-distribution directions along the subintercardinals, e.g., North-northeast, test whether $\bfg$ lies on a roughly circular manifold.
Each trial is a rollout of a policy over $\qty{60}{\second}$ and policies are always compared with the same initial conditions and seed.
Double integrator robots are initialized uniformly on the disk of radius $\sqrt{4 N / \pi}$ with random initial speeds in $\pm \qty{5}{\meter\per\second}$, stepped at $dt=\qty{0.1}{\second}$ with zero acceleration.

Policies use Qwen3-1.7B as the default reasoning engine with sampling temperature $0.6$.
Each robot receives a unique thinking trace, $\bfT$ of length $160$, generated by the same operator prompt $\bfP$ unless otherwise specified.
This emulates the operator voicing a single goal to the team, but each robot reasoning about it individually.
We found that the length of the thinking trace can be thought of as another hyperparameter and $160$ is an empirically determined sweet-spot (we validated 120 and 320 as well).
The feedback token is refreshed using MAST output at a rate of $\qty{1}{\hertz}$, while the inner MAST control loop runs at $\qty{10}{\hertz}$.
All initial positions and thinking traces are held-out from training and are generated for evaluation.

We compare COMPASS to the oracle expert it was trained on, a frontier model LLM serving as a centralized controller, and two levels of ablation: (i) COMPASS without communications, i.e., $r_c = 0$.
(ii) A frozen SLM with learned register reasoning state readout of comparable size to MAST, which encodes neighboring state in language, e.g., ``...My state: pos (x, y), velocity (vx, vy), Neighbor1: relative
  pos (x, y), velocity (vx, vy), Neighbor 2 …, Average neighbor vel (vx, vy)''.

The heading error is computed by measuring the heading of the center of mass, $\theta_\mathrm{COM}$ and comparing it to the ground truth heading, $\hat{\theta}$ of the given directive, $e_h = \vert (\theta_\mathrm{COM} - \hat{\theta} + 180) \bmod 360  - 180 \vert$.
The fragmentation number describes whether the swarm is connected ($\mathrm{frags} = 1$) or not ($\mathrm{frags} > 1$).
Velocity variance is the variance in the agents's velocities about the fleet mean over the settled window; it measures the consensus of the swarm.
The coefficient of variation (CV) of the robots' nearest-neighbor distances summarizes the spatial uniformity of the system.
Overly clumped flocks inflate the CV while an even lattice drives it to zero.

\subsection{Scalability}\label{sec:res_scale}
\begin{figure*}[htp!]
    \centering
    \includegraphics[width=\linewidth]{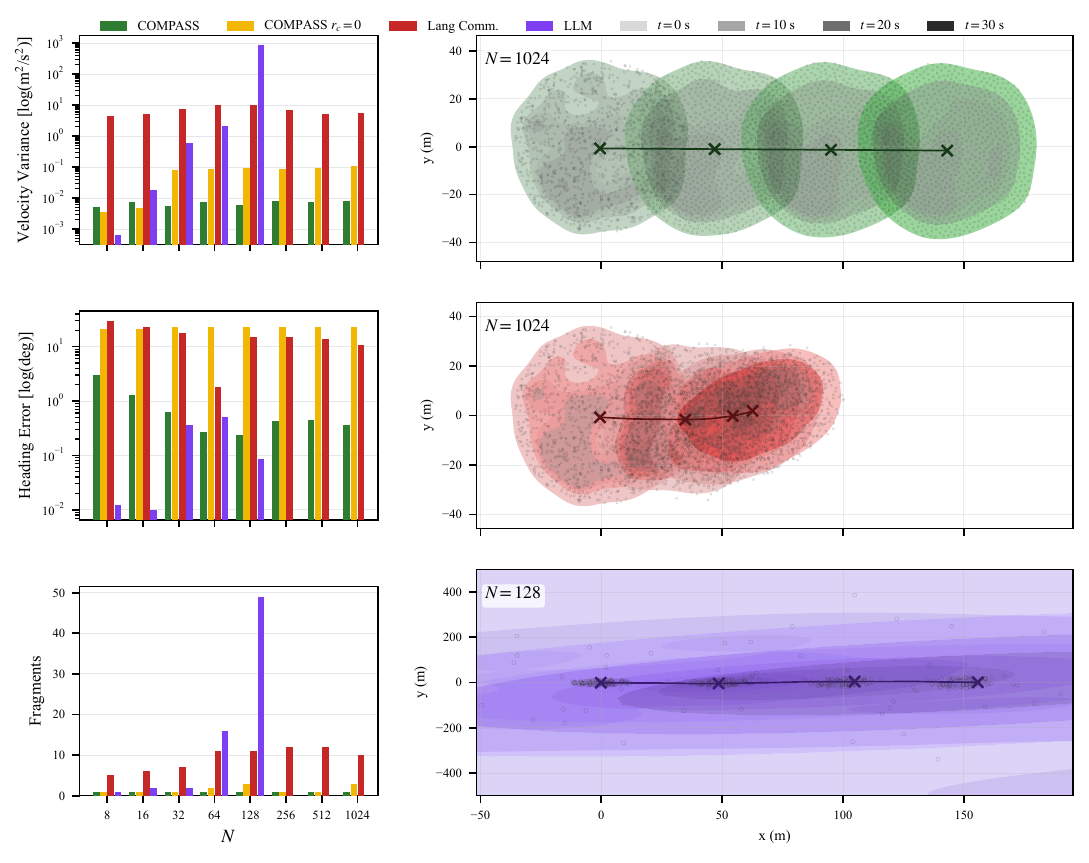}
    \caption{
        Scalability evaluation for the prompt ``go east'' for $N = [8, 16, 32, 64, 128, 256, 512, 1024]$ and $\qty{30}{\second}$ rollouts.
        Our model, COMPASS (green) is compared against a centralized controller with hand engineered feedback using API calls to Claude Opus 4.8 (purple) \cite{ClaudeOpus48} and the ablated language-only communication method (red).
        \textbf{Left:} Velocity variance (consensus) and heading error are shown in log-scale and fragmentation measured during the last $\qty{15}{\second}$ of the rollout at each $N$.
        \textbf{Right:} Rollouts for each policy shown at the largest evaluated scale.
        The trajectory of the center of mass (COM) overlays the migrating flock, marked with an $\times$ at each of the captured snapshots in time.
    }
    \label{fig:scale}
\end{figure*}
Although we train COMPASS on $N=[32, 64]$, we evaluate the system and baselines on scales of up to $N=1024$ to determine how the system performs under increasingly challenging settings.
The results in \fgref{fig:scale} illustrate that COMPASS maintains consistent performance across scale and heading error \emph{decreases} with scale.
This is a boon of collective reasoning -- the consensus of many diverse reasoning outputs, integrated through the multi-hop communication of MAST, provides the swarm with a more accurate heading.
The error goes from $2.79^\circ$ at $N=8$ to $0.34^\circ$ at $N=1024$; scale brings tangible advantages.

Under language-only communication, neighboring states are front-loaded into the semantic portion of a robot's prompt and the $N=64$ flock fragments into $15$ groups and slows down, traveling at an average velocity of $\qty{2.5}{\meter\per\second}$ (half of the target velocity).
This degenerate result is worse than ablating communication altogether and appears to stem from requiring the thinking trace to reason over substantial spatial information from neighboring agents.
This is a task that even frontier models struggle with when not aided by tooling.
Diagnosing the thinking trace reveals that reasoning over the neighbor state overwhelms the model and little compute is spent determining the intended direction of heading.

\begin{figure*}
    \centering
    \includegraphics[width=\linewidth]{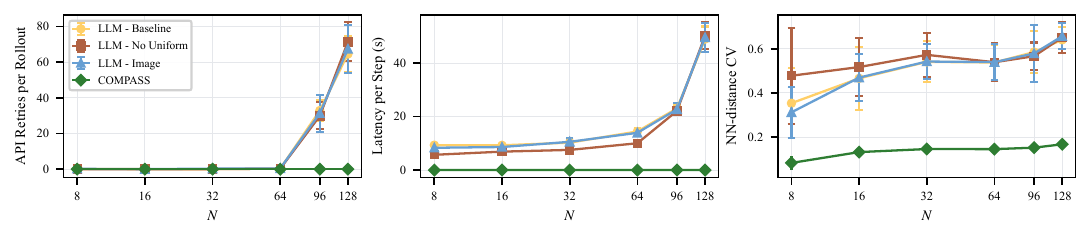}
    \caption{Performance of Claude Opus 4.8 tasked to flock a team of robots over growing scales in a centralized scenario.
        The baseline prompt (yellow) requested flocking with uniform distance between agents. 
        The no-uniform prompt (brown) limited the task to .
        The image prompt (blue) incorporates vision for flocking with uniform distance.
        We evaluate COMPASS on the same initial conditions and demonstrate that it maintains near-uniform nearest-neighbor spacing and latency as the fleet grows.
        We use the coefficient of variation of each robot's nearest-neighbor distance, averaged over the rollout, to evaluate the spatial distribution of the fleet across scales.
    }
    \label{fig:llms-dont-scale}
\end{figure*}

A centralized LLM-based controller is able to decode the given semantic directive perfectly, flying the core of the flock in the correct direction.
As $N$ increases, the LLM fails to provide coherent accelerations to all robots and the flock fragments as more of these robots execute ill-coordinated accelerations.
Sweeping across scales (\fgref{fig:llms-dont-scale}), we find that the centralized LLM demonstrates two failure modes: (i) generating accelerations that satisfy both consensus and cohesion and (ii) the generation of the incorrect \emph{number} of actions for the team that becomes worse with increasing $N$ and requires a retry to obtain the correct number of actions; \fgref{fig:llms-dont-scale}.
This highlights the scalability failure of a centralized LLM controller, which COMPASS resolves natively.
We remark that, given enough compute on each robot, the frontier model itself could serve as the LM in COMPASS.

\subsection{Structured Diversity}\label{sec:res_collective}
\begin{figure*}[htb]
    \centering
    \includegraphics[width=0.9\linewidth]{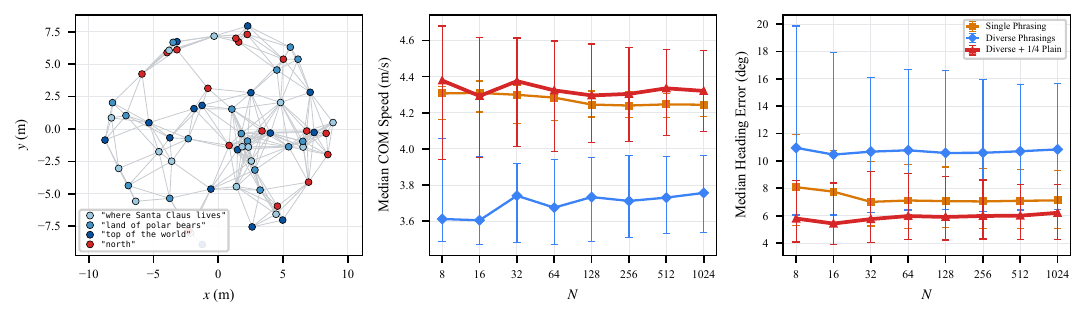}
    \caption{
        \textbf{Left:} Example initialization where four different phrasings are distributed over the team. A 1/4 of the team receives a simpler command for the LM to reason over.
        \textbf{Middle:} Median COM speed averaged over the trajectory against team size $N$ on a selection of cardinal vague prompts.
        \textbf{Right:} Median settled heading error over $N$.
    }
    \label{fig:mixed_phrasing}
\end{figure*}
Does a fleet of LMs reason about a vague command better than one?
We find that the agentic crowd is unwise when given same prompt or when each robot receives a different vague phrasing for the same directional intent, i.e., robot $i$ receives phrasing $i \bmod 4$; \fgref{fig:mixed_phrasing}.
COMPASS is able to leverage an advantage inherent to a collective of LMs: it can provide diverse re-phrasing of the task; crucial when some phrasings are easier to reason over than others.
Phrase diversity with an included easier rephrasing of the task is a boon that enables recovery of the intent through feedback via MAST; both in terms of heading error and speed of the flock.
Notably, the gaps in capabilities remain consistent across scales, evaluated over 10 trials at each scale. 

\subsection{Command Generalization}\label{sec:res_lang}
COMPASS embeds an off-the-shelf LM to serve as a reasoning engine.
A key advantage of COMPASS is the interchangeability of the LM, providing deployment flexibility for platform and task constraints.
We measure the heading error and latency of three sizes of the Qwen3 LM on our cardinal navigation task, trained with an MLP instead of MAST and evaluated as individual agents ($N=1$); \tabref{tab:engines}.
The linear head regresses $\bfz_i(t)$ onto the ground-truth unit bearing, $(\cos\theta, \sin\theta)$. 
Reasoning state $\bfz_i(t)$ is read from the last layer of each model for compatibility of comparison; empirically we find that extracting $\bfz_i(t)$ from a deeper layer improves performance on vague inputs but the relationship between model layer and accuracy is non-linear and this comparison is designed to be relative.
A layer-sweep over the selected model enables the optimal layer selection for the given task.
\begin{table}[htp!]
    \centering
    \caption{
    Reasoning-engine tradeoff on zero-shot vague decoding.
    }
    \label{tab:engines}
    \begin{tabular}{lcccc}
      \toprule
      Engine & $\ell$ & $d_{\mathrm{LM}}$ & Err. (v. ref) & Latency / VRAM\\
      \midrule
      Qwen3-0.6B (ref) & 28 & 1024 & 1.00 & 3.7\,s / 1.2\,GB \\ 
      \textbf{Qwen3-1.7B} & 28 & 2048 & 0.80 & 4.0\,s / 3.3\,GB \\ 
      Qwen3-4B   & 36 & 2560 & 0.65 & 5.0\,s / 7.7\,GB \\ 
      \bottomrule
    \end{tabular}
\end{table}

Larger models gain improved heading on the held-out vague prompts at the cost of higher latency and compute required, measured on an RTX A5000; \tabref{tab:engines}.
The simplified probe is trained until each LM obtains a fit of approximately $5^\circ$ on the training set and is therefore only for comparison and not representative of the final performance in COMPASS.
We select Qwen3-1.7B with reasoning readout at $\ell=27$ as the default configuration for evaluation as it is edge-deployable while generalizing well to unseen vague operator inputs.

\subsubsection{Vague Commands}
\begin{figure}
    \centering
    \includegraphics[width=\linewidth]{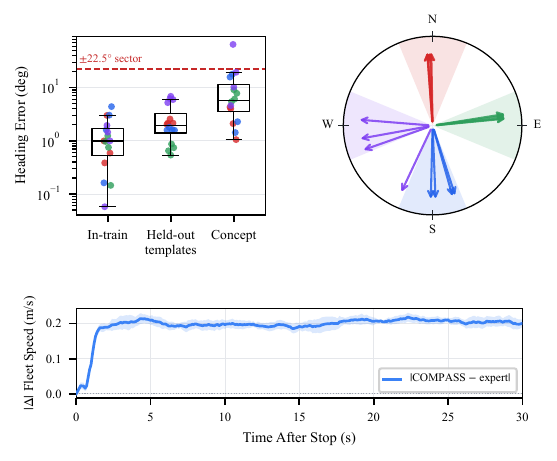}
    \caption{
    Zero-shot generalization to new prompt classes ($N=64$, cardinal directions).
    \textbf{Left:} The box plot shows the settled heading error per evaluation category; strip plot points are colored by the commanded direction.
    Each class has $16$ flights: $4$ per cardinal direction.
    \textbf{Right:} the settled headings for the held-out concept commands; arrows are colored according to the commanded direction.
    \textbf{Bottom:} The COMPASS flies the command ``go east'' for $\qty{15}{\second}$ and then the command switches to ``stop'' with thinking regenerated at the switch.
    The trace shows the absolute difference $|$ COMPASS $-$ expert $|$ in mean fleet speed over time and over $10$ trials with $N=64$.
    }
    \label{fig:compass_across_classes}
\end{figure}

Piloting swarms of $N=64$ robots, we sample both in-distribution and held-out prompt templates, e.g., ``new heading \{surface\}'' as well as different prompt classes; \tabref{tab:corpus}.
These episodes show that COMPASS is able to generalize zero-shot to held-out prompts, \fgref{fig:compass_across_classes}, while simultaneously maintaining a cohesive flock.
Generalizability is a property driven by the reasoning engine.
The thinking traces often reveal that smaller models sometimes struggle to select a direction for a vague prompt.
For $16$ trials, the median heading error per class is: in-training $1.0^\circ$, held-out templates $1.9^\circ$, concept $5.7^\circ$.
When the fleet is commanded to stop mid-mission, it retains a small drift velocity of $\qty{0.2}{\meter\per\second}$ compared to the virtually stationary expert.

\subsubsection{Unseen Headings}
\begin{figure}[!tb]
    \centering
    \includegraphics[width=\linewidth]{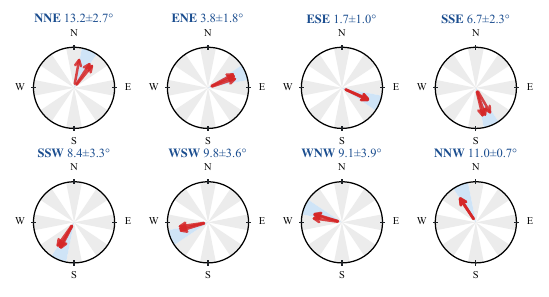}
    \caption{
        Settled COM headings (red) for the eight held-out secondary intercardinals, $N=64$, $\qty{60}{\second}$ rollouts.
        The shaded sector is the commanded 16-wind sector; each panel reports its mean $\pm$ std error.
    }
    \label{fig:bearing_polar}
\end{figure}

For natural language commands, COMPASS is trained on the 8-point compass rose and Stop; does the reasoning state $\bfz_i(t)$ generalize to new directions zero-shot?
We evaluate this by adjusting the wrapper to request reasoning for the 16-point compass rose, which includes the secondary intercardinal directions such as North-northeast, which bisect the $45^\circ$ sectors at $22.5^\circ$ angles; \fgref{fig:bearing_polar} and \tabref{tab:bearing}.
Despite being trained on the 8-point rose, the policy generalizes to a 16-point rose without any retraining.
The 16-point wrapper itself is not included in the training examples either.
COMPASS achieves a mean error of $7.96^\circ \pm 4.41^\circ$ on secondary intercardinals for over $24$ trials.
While the zero-shot performance degrades, it also hints that $\bfz_i(t)$ is able to generate outputs for new angles that are better than random directions or the nearest direction on the 8-point rose.
The policy does not generalize equally well to every held-out secondary intercardinal direction and has the lowest mean heading error when ``east'' is given as the first direction, $1.7 \pm 1.0^\circ$ and $3.8 \pm 1.8^\circ$ for ESE and ENE, respectively.
Direction NNE is the worst offender, falling outside the sector in 2/3 trials with an average error of $13.2 \pm 2.7^{\circ}$; overall 18/24 trials have in-sector headings.

COMPASS is trained on $720$ numerically expressed bearings wrapped in natural language, with North at $0^\circ / 360^\circ$ (aeronautical convention) and clockwise increments, COMPASS is able to ground bearings in $\bfz_i(t)$ such that out-of-distribution bearings in prompts are flown with mean error $2.2 \pm 1.6^\circ$; \tabref{tab:bearing}.
Empirically, we find that training on bearings did not improve the performance on the secondary intercardinals, even if the equivalent bearings are included in the training set.
Therefore, grounding of new directions in language is independent of numeric bearings.
\begin{table}[!tb]
    \centering
    \caption{
        Generalization to the 16-point rose and numeric bearings.
    }
    \label{tab:bearing}
    \begin{tabular}{lccc}
        \toprule
        Direction family & Trials & In sector & Med.\ err.\ [$^\circ$] \\
        \midrule
        Cardinals (trained) & 20 & 100\% & 1.1 \\
        Intercardinals (trained) & 20 & 100\% & 1.2 \\
        Sec.\ intercardinals (zero-shot) & 24 & 75\% & 7.7 \\
        Numeric bearings (zero-shot) & 16 & N/A & 2.0 \\
        \bottomrule
    \end{tabular}
\end{table}

%% file: sections/06_conclusion.tex
We introduced COMPASS, a decentralized, scalable architecture designed to elicit and harness the collective dynamics of an agentic fleet of robots.
Our experiments elucidate the architectural roles: the language channel decides \emph{where to go} and is limited by the biases of the chosen LM.
Engineered methods such as structured diversity can be employed to limit that bias, demonstrating performance gains that hold across scales.
MAST decides \emph{getting there cohesively}, integrating agentic decisions into motor commands through multi-hop communication at variable $N$.
The reasoning state acts as the downstream interface between the components; the language model never needs to reason spatially.
A learned feedback token enables collective information diffused through MAST to appear in the input of the LM for the next reasoning step.

With language steered flocking as the backdrop, we studied how scale and feedback shape the system's performance.
We found that scale synergizes well with language under \emph{structured} diversity and that the wisdom of the crowd is not guaranteed.
A diverse mixture of phrasings of similar decoding difficulty does not improve performance, while mixing in an easy phrasing across a quarter of the fleet can promote bias cancellation.
Feedback is most effective when it combines a learned representation that the language model may reason over.
When introducing feedback into the language channel, small language models that are edge-deployable on SWaP-limited hardware can easily be overwhelmed by too much feedback, i.e., thinking becomes re-tasked.
In principle, the COMPASS loop is not task specific and tasks that are verifiable and have a language-expressible objective can operate the same loop.